\documentclass[sigconf]{acmart}

\usepackage[most]{tcolorbox}
\usepackage[normalem]{ulem}
\usepackage{tabularray}
\usepackage{makecell} 
\usepackage[normalem]{ulem}
\useunder{\uline}{\ul}{}
\usepackage{algorithm}
\usepackage{algorithmic}
\usepackage{multirow}
\usepackage{amsmath} 
\usepackage{enumitem}
\usepackage{balance}
\usepackage{caption} % 载入caption宏包
\usepackage{array} 
\usepackage{etoolbox} % 载入etoolbox宏包
\usepackage{graphicx}
\usepackage{booktabs}
\setcopyright{acmcopyright}

\copyrightyear{2025}
\acmYear{2025}
\setcopyright{cc}
\setcctype{by}
\acmConference[SIGIR '25]{Proceedings of the 48th International ACM SIGIR Conference on Research and Development in Information Retrieval}{July 13--18, 2025}{Padua, Italy}
\acmBooktitle{Proceedings of the 48th International ACM SIGIR Conference on Research and Development in Information Retrieval (SIGIR '25), July 13--18, 2025, Padua, Italy}\acmDOI{10.1145/3726302.3730295}
\acmISBN{979-8-4007-1592-1/2025/07}

\begin{document}
\title{JuDGE: Benchmarking Judgment Document Generation for Chinese Legal System} 
\author{Weihang Su}
\email{swh22@mails.tsinghua.edu.cn}
% \affiliation{Department of Computer Science and Technology, Tsinghua University}
\affiliation{
DCST, Tsinghua University\\
Beijing 100084 \country{China}
}

\author{Baoqing Yue}
\affiliation{
DCST, Tsinghua University\\
Beijing 100084 \country{China}
}

\author{Qingyao Ai}
\authornote{Corresponding author}
\email{aiqy@tsinghua.edu.cn}
\affiliation{
DCST, Tsinghua University\\
Beijing 100084 \country{China}
}

\author{Yiran Hu}
\affiliation{
DCST, Tsinghua University\\
Beijing 100084 \country{China}
}

\author{Jiaqi Li}
\affiliation{
DCST, Tsinghua University\\
Beijing 100084 \country{China}
}

\author{Changyue Wang}
\affiliation{
DCST, Tsinghua University\\
Beijing 100084 \country{China}
}

\author{Kaiyuan Zhang}
\affiliation{
DCST, Tsinghua University\\
Beijing 100084 \country{China}
}

\author{Yueyue Wu{\fontsize{9pt}{10pt}\selectfont {*}}}
% \authornote{Corresponding author}
\email{wuyueyue1600@gmail.com}
\affiliation{
DCST, Tsinghua University\\
Beijing 100084 \country{China}
}

\author{Yiqun Liu}
\affiliation{
DCST, Tsinghua University\\
Beijing 100084 \country{China}
}

\renewcommand{\shortauthors}{Su, et al.}

\begin{abstract}

This paper introduces \textbf{JuDGE} (\underline{\textbf{Ju}}dgment \underline{\textbf{D}}ocument \underline{\textbf{G}}eneration \underline{\textbf{E}}valuation), a novel benchmark for evaluating the performance of judgment document generation in the Chinese legal system.
We define the task as generating a complete legal judgment document from the given factual description of the case.
To facilitate this benchmark, we construct a comprehensive dataset consisting of factual descriptions from real legal cases, paired with their corresponding full judgment documents, which serve as the ground truth for evaluating the quality of generated documents.
This dataset is further augmented by two external legal corpora that provide additional legal knowledge for the task: one comprising statutes and regulations, and the other consisting of a large collection of past judgment documents.
In collaboration with legal professionals, we establish a comprehensive automated evaluation framework to assess the quality of generated judgment documents across various dimensions.
We evaluate various baseline approaches, including few-shot in-context learning, fine-tuning, and a multi-source retrieval-augmented generation (RAG) approach, using both general and legal-domain LLMs. 
The experimental results demonstrate that, while RAG approaches can effectively improve performance in this task, there is still substantial room for further improvement\footnote{All the codes and datasets are available at: https://github.com/oneal2000/JuDGE}.

\end{abstract}




\maketitle

\section{Introduction}

In recent years, the rapid development of large language models (LLMs) has enabled numerous applications in the legal domain~\cite{brown2020language,touvron2023llama,scao2022bloom,yang2024qwen2}, including contract analysis, case law research, and legal document drafting~\cite{shu2024lawllm,yue2023disc,wang2024lekube,su2023legalaid}.
Among these applications, judgment document generation stands out as a particularly promising yet challenging task. 
Judgment documents are authoritative court records that encapsulate legal reasoning, procedural details, and final rulings~\cite{kumar2013finding,li2023sailer}. 
As essential components of judicial proceedings, these documents serve as authoritative records of legal decisions and reflect the complexity of legal reasoning and statutory interpretation~\cite{zander2015law}.
In practice, drafting judgment documents is a highly labor-intensive task that consumes substantial time and resources in courts.
Specifically, creating a judgment document requires the judge to gather a large amount of legal information, including relevant statutes, past case precedents, and fundamental legal knowledge.
Once the relevant legal information has been collected, it must be systematically organized and integrated with professional legal reasoning to generate a well-structured and legally sound judgment document.
Due to the time-consuming and labor-intensive nature of drafting judgment documents, AI technologies hold great promise for supporting legal professionals.
For example, advanced IR techniques can systematically collect and organize relevant legal information, while generative AI can support drafting well-structured legal texts.

Nonetheless, despite the rapid development of LLMs, their application to judgment document generation remains largely unexplored. 
This gap arises primarily due to two critical factors.
Firstly, judgment document generation is a complex domain-specific task that requires extensive legal expertise and knowledge. 
General-purpose LLMs often lack the specialized legal knowledge necessary to conduct complex legal reasoning and produce high-quality legal texts. 
More importantly, the absence of standardized benchmarks and automated evaluation methods significantly hinders progress in this field.
Currently, the quality of automatically generated judgment documents can only be assessed through expert annotation, which is both time-consuming and unscalable.
Consequently, establishing robust benchmarks and automated evaluation frameworks is crucial to overcoming the bottleneck in this field.

To address these challenges and fill this research gap, we introduce \textbf{JuDGE} (\underline{\textbf{Ju}}dgment \underline{\textbf{D}}ocument \underline{\textbf{G}}eneration \underline{\textbf{E}}valuation), a novel benchmark for evaluating the performance of judgment document generation. In collaboration with legal professionals, we further establish a comprehensive, automated evaluation framework spanning four dimensions: {penalty accuracy}, {convicting accuracy}, {referencing accuracy}, and {documenting similarity with the ground truth}. 
Crucially, JuDGE is designed to mirror real-world judicial reasoning, which typically follows a systematic syllogistic approach: the major premise comprises relevant statutes, precedents, and jurisprudence, while the minor premise is drawn from the factual circumstances of the case. 
In practice, judges typically begin by establishing the major premise comprising relevant statutes, precedents, and jurisprudence. Then summarize the minor premise from the case’s factual circumstances. By synthesizing these elements through legal reasoning, they conclude the judgment and draft the judgment document detailing both the reasoning process and the final decision.
To reflect this real-world process, the JuDGE benchmark focus on generating the full judgment documents based on the factual descriptions, which are sufficient as the minor premise for judgment document generation.
Specifically, the dataset incorporates two specialized legal corpora (covering statutes, precedents, and other domain-specific texts) to serve as the major premise, alongside a set of publicly available factual descriptions from official sources that function as the minor premise. 
We also provide the corresponding ground-truth full judgment documents for each instance, offering the reference for evaluation.

To demonstrate the use of our benchmark and facilitate future research, we evaluate multiple baseline approaches using a variety of LLMs. 
The baselines include standard methods that directly employ general-purpose and legal-domain LLMs to generate complete judgment documents based on the provided case facts. 
Moreover, we investigate more advanced strategies that leverage few-shot in-context learning and supervised fine-tuning (SFT) to enhance the models’ performance on this task. 
Beyond these conventional approaches, we propose a more robust baseline based on multi-source retrieval-augmented generation (Multi-source RAG), which incorporates external knowledge from both statute corpus and judgment document corpus. 
This strong baseline is intended to serve as a reference point for future work on this task.

Experimental results on the Judgment Document Generation task show that both general and domain-specific legal LLMs struggle to produce high-quality legal documents when used directly or with in-context learning. 
Augmenting these models via fine-tuning and Multi-source RAG leads to performance improvements, but there is still considerable room for improvement, underscoring the challenging nature of this task.
These findings highlight the need for further exploration in this task and innovation at the intersection of advanced information retrieval techniques and generative AI.

In summary, our contributions are threefold:
\begin{enumerate}[leftmargin=*]

\item We introduce JuDGE, a novel benchmark tailored for judgment documents generation, incorporating a comprehensive dataset designed to reflect real-world judicial reasoning processes.

\item We establish a robust automated evaluation framework that systematically evaluates the quality of generated judgment documents across multiple dimensions, facilitating scalable and systematic evaluation for this task.

\item We conduct experiments and analyses of various baseline approaches using different LLMs. Our findings serve as valuable references for future research, while also illuminating key challenges and opportunities for advancement in this task.

\end{enumerate}

\section{Preliminaries and Task Definition}

This section introduces the foundational concepts and formal definitions of the Judgment Document Generation task.
% This section provides an overview of the foundational concepts and the formal task definitions for the Judgment Document Generation task. 
% We introduce the structure and notation of judgment documents and formalize the task of generating judgment documents from fact descriptions.

\subsection{Judgment Document Structure and Notation}

\begin{figure}[t]
\centering
    \includegraphics[width=\columnwidth]{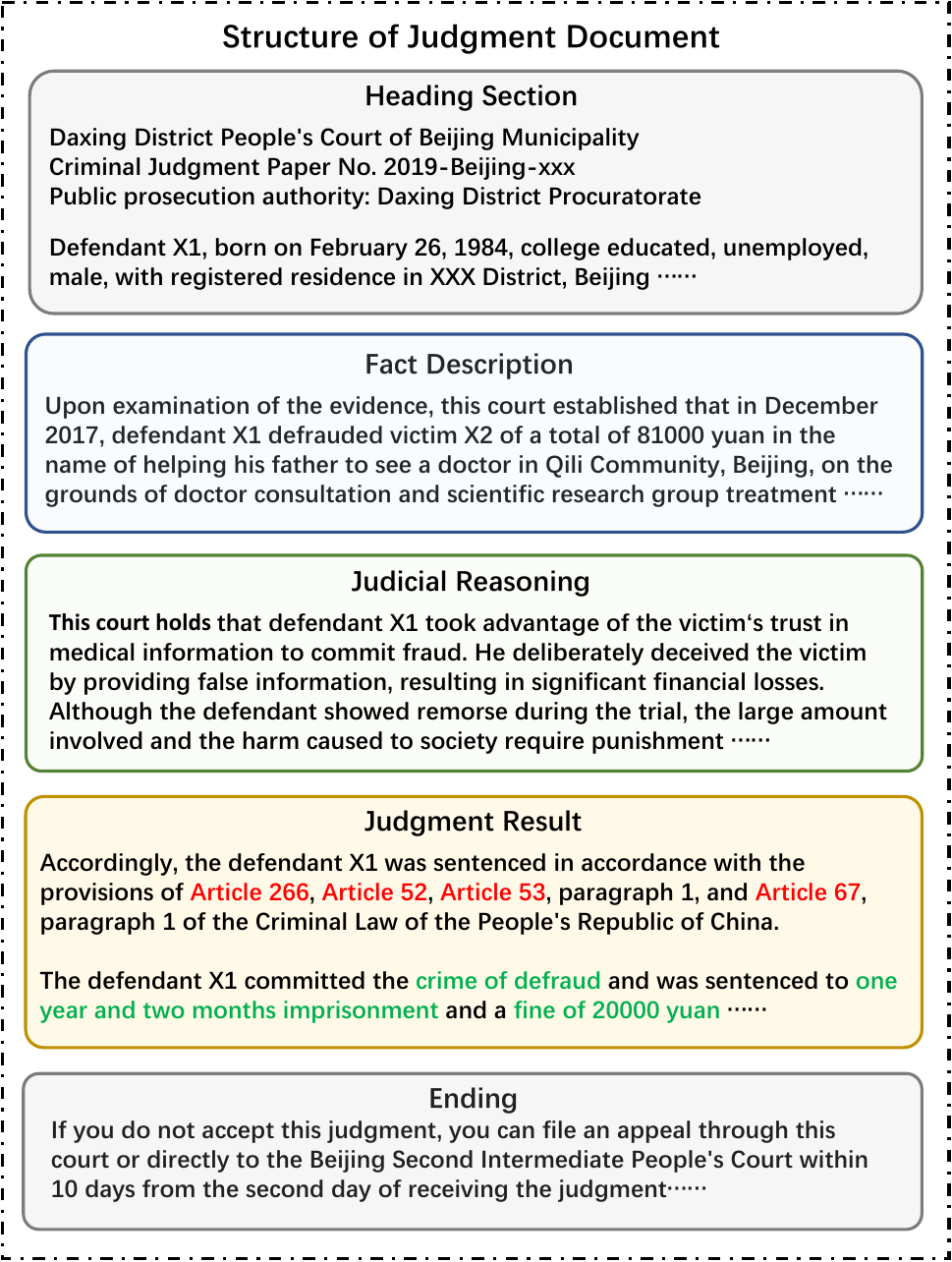}
    \vspace{-2mm}
    \caption{An illustration of the structure of a judgment document.} 
    \label{fig:1}
    \vspace{-3mm}
    \label{pic:framework}
\end{figure}

Let $\mathcal{F}$ denote the space of fact descriptions extracted from judgment documents. 
Each fact description $f \in \mathcal{F}$ provides a clear and unbiased summary of the circumstances underlying a legal dispute. 
The corresponding judgment document (denoted by $j$) belongs to the space $\mathcal{J}$. 
% A complete judgment document typically comprises several interdependent sections (illustated in Figure \ref{fig:1}), including but not limited to:
A complete judgment document typically consists of several interdependent sections, as illustrated in Figure \ref{fig:1}, including but not limited to:

\begin{itemize}[leftmargin=*]
    \item \textbf{Heading Section}: 
    This section provides essential administrative details that establish the jurisdiction and authenticity of the document. 
    It includes the court's name, the case number, the presiding judge's name, and the date of judgment. 
    Defendant information, such as full name, gender, date of birth, and nationality, is also presented. 
    These details formally identify the parties involved and set the context for the proceedings.
    
    \item \textbf{Fact Description}: 
    This section offers a clear and comprehensive narrative of the events and circumstances that led to the legal dispute. 
    It outlines the key facts and actions in a concise and objective manner. 
    The description is crucial for establishing the factual basis of the case. 
    It provides the necessary context for understanding the subsequent legal analysis.
    
    \item \textbf{Judicial Reasoning}: 
    This section details the court's evaluation of both factual and legal aspects of the case.
    It systematically applies relevant laws, legal principles, and judicial precedents to the facts. 
    The analysis discusses the evidence, the offense's nature, and the liability's determination. 
    It ensures that the legal reasoning behind the decision is transparent and well-founded.
    
    \item \textbf{Judgment Result}: 
    This final section presents the court's conclusive decision along with its legal justification. 
    It explicitly cites the applicable statutes, regulations, and legal provisions that support the ruling. 
    Additionally, the section outlines the specific charges, sentencing details (including the prison term and fines), and any other imposed penalties.
    % It succinctly summarizes the outcome and its legal basis.
\end{itemize}

The interdependence between these sections ensures the judgment document maintains logical consistency. 
Each section builds upon the previous one, with the \textbf{Judicial Analysis Process} being generated based on the \textbf{Fact Description} while simultaneously providing the legal foundation for the \textbf{Judgment Result}.

\subsection{Task Definition}
\label{sec:task_definition}

The task of \emph{Judgment Document Generation} is formalized as a conditional text generation problem. 
Given an input fact description $f \in \mathcal{F}$, the objective is to generate a complete judgment document $\hat{j} \in \mathcal{J}$ that is both structurally coherent and legally sound. 
Formally, the goal is to learn a mapping function:

\begin{equation}
    \mathcal{M} : \mathcal{F} \rightarrow \mathcal{J},
\end{equation}

\noindent such that for a given $f$, the generated document $\hat{j} = \mathcal{M}(f)$ approximates the ground truth $j$ in terms of content, structure, and legal validity. 
\textbf{In real-world applications, this mapping function is implemented through a domain-specific LLM or an automated system.}

In summary, the task of Judgment Document Generation involves developing an automatic system that transforms a given fact description $f \in \mathcal{F}$ into a complete and legally valid judgment document $\hat{j} \in \mathcal{J}$. 
Using our dataset, the fact description serves as input to the system, and the output is the generated full judgment document. 
The system’s performance is then automatically evaluated by comparing the generated document against the ground truth across four perspectives, using twelve distinct metrics, as detailed in Section \S \ref{sec:evaluation}.

% By leveraging both the inherent structure of legal texts and external legal knowledge sources ($K_L$ and $K_J$), the proposed system aims to facilitate the drafting process and significantly reduce the workload on legal professionals.

\section{Dataset Construction}
This section outlines the construction of the JuDGE dataset, starting with the collection of criminal case documents and the statute corpus. 
Next, it describes the pre-processing steps to ensure high data quality, followed by expert annotations to verify the dataset’s accuracy and consistency. 
The section then presents key dataset statistics, offering insights into its composition, and concludes with a discussion of the ethical considerations, highlighting our commitment to privacy, transparency, and regular updates.

\subsection{Data Collection}

To construct the judgment document corpus for our dataset, we collected publicly available criminal case documents from the China Judgments Online platform, which is maintained by the Supreme People’s Court of China, in line with the approach taken by \citeauthor{ma2021lecard}~\cite{ma2021lecard}. 
These documents span a 20-year period and were randomly selected from an extensive database containing approximately six million legal cases. 
The selection process ensured that the corpus included a diverse range of criminal cases, covering various legal principles and judicial outcomes.

To construct the statutory articles corpus that serves as external knowledge for our model, our legal team conducted detailed annotations to identify the most relevant and up-to-date Chinese statutory laws and regulations. 
These statutes were manually downloaded from official government websites to ensure accuracy and currency.
The laws were then segmented into individual articles using automated scripts, facilitating easy retrieval and application for downstream tasks. 
The complete list of statutes selected for inclusion is publicly available on our official GitHub repository\footnote{https://github.com/oneal2000/JuDGE}.

\subsection{Data Pre-processing}
\label{sec:data-pre}

In the pre-processing phase, we implemented key filtering criteria to ensure the quality and consistency of the dataset. The primary filtering mechanisms are as follows:
Firstly, we applied length restrictions to maintain document consistency with typical legal texts and ensure manageable input sizes for model processing. Specifically, the fact description of each judgment document is limited to a maximum of 1,000 Chinese characters, while the full judgment text ranges from 1,000 to 3,000 Chinese characters.
Secondly, we filtered out case documents that don't have key legal elements, such as the Criminal Law articles, the crime type, and the sentence duration, which are critical for establishing the legal validity of the judgment. 
We also required the inclusion of standard legal phrases like “this court believes” and “the judgment is as follows” to ensure adherence to legal language conventions.
Specifically, we designed regular expressions to extract and validate each section\footnote{All the specific code implementations related to Data Pre-processing are available at: https://github.com/oneal2000/JuDGE}. For example, the opening of the Fact Description section typically starts with phrases such as ``This court finds'' or ``It has been established through trial''. 
The regular expressions can cover 98\% of documents in a dataset consisting of millions of judgment documents, with the remaining 2\% being non-standard or improperly formatted.

After filtering, we transform the Judgment Documents into structured representations, represented as a series of (key, value) pairs.
The specific fields for each document include CaseID, Fact Description Section, Judicial Reasoning Section, Judgment Result Section, Sentence Length, Fine, Crime Type, and Referenced Legal Articles. 
These fields are summarized in Table \ref{tab:data_fields}, which presents the keys and their descriptions.

\begin{table}[t]
\centering
\caption{Fields and Descriptions of the JuDGE. This table lists the key fields included in the dataset, along with their respective descriptions.}
\vspace{-2mm}
\label{tab:data_fields}
\setlength\tabcolsep{1.8pt}
{\small
\begin{tabular}{lc}
\toprule
\textbf{Key} & \textbf{Description} \\
\midrule
\textbf{CaseID} & Unique identifier for the case \\
\textbf{Fact} & Section summarizing the key facts of the case \\
\textbf{Reasoning} & Section detailing the legal reasoning behind the judgment \\
\textbf{Judgment} & Section outlining the court’s final decision and penalties \\
\textbf{Sentence} & Duration of the prison sentence \\
\textbf{Fine} & Monetary penalty imposed \\
\textbf{Crime Type} & The type of crime involved in the case \\
\textbf{Law Articles} & Legal statutes cited in the judgment \\
\bottomrule
\end{tabular}
}
\vspace{-5mm}
\end{table}

\subsection{Expert Annotation}

While the data pre-processing steps provided an initial quality filter for judgment documents, we further enhanced the JuDGE dataset through expert annotation.
To achieve this, we recruited law students to evaluate each judgment document instance based on five critical aspects: document formatting, the correctness of judicial reasoning, appropriate legal references, the accuracy of crime classification, and the reasonableness of the sentencing. 
Each document was annotated by two independent experts. 
If both annotators agreed that the document met the quality criteria, it was classified as “usable.” If discrepancies arose between annotators, the document was excluded from the dataset.
The annotation team consisted of seven members recruited from prominent law schools. 
Our compensation plan, which was based on working hours, ensured that the annotation process was incentivized fairly, offering an average hourly wage of 45 CNY, which is significantly higher than the minimum wage requirements in Beijing. 
To evaluate the reliability of the annotations, we applied Cohen’s Kappa coefficient in a binary classification context. 
The analysis, performed on 2,574 annotated instances, yielded a Kappa value of 0.8012, indicating very high inter-annotator agreement. 
This strong agreement reflects the consistency and reliability of the expert annotations.
After applying these filtering standards and expert annotations, we obtained the final dataset with 2,505 legal documents covering 182 different statutes and 142 distinct crime types. 
This selection process ensured the dataset’s diversity in legal references and crime types while also maintaining its high quality for the benchmark.

\subsection{Dataset Statistics}

\begin{table}[t]
\centering
\caption{Basic Statistics of the JuDGE Dataset. ``Total Fact-Judgment Pairs'' indicates the total number of paired entries in the dataset; ``Unique Charges'' and ``Unique Criminal Law Provisions'' refer to the number of distinct charges and law articles covered in the test set. All length metrics are measured in Chinese characters.}
\vspace{-2mm}
\label{tab:basic_statistics}
\begin{tabular}{lc}
\toprule
\textbf{Statistic} & \textbf{Number} \\
\midrule
Total Fact-Judgment Pairs & 2,505 \\
Training Set Size & 2,004 \\
Test Set Size & 501 \\
Unique Charges & 142 \\
Unique Criminal Law Provisions & 182 \\
Avg. Fact Length & 651.95 \\
Avg. Reasoning Length & 281.75 \\
Avg. Judgment Result Length & 207.06 \\
Avg. Full Document Length & 1,741.56 \\
Avg. Charges per Document & 1.26 \\
Avg. Statutory Articles per Document & 4.31 \\
\midrule
External Judgment Documents Corpus Size & 103,251 \\
External Statutory Articles Corpus Size & 55,348 \\
\bottomrule
\end{tabular}
\end{table}

Table~\ref{tab:basic_statistics} summarizes the core numerical attributes of the JuDGE dataset and its supplementary legal corpora. 
The dataset contains 2,505 fact-judgment pairs, covering a diverse range of criminal cases with 142 unique charges and 182 distinct criminal law provisions. 
This diversity suggests that the dataset is highly applicable for modeling various legal scenarios.
The average lengths of the different sections further reflect the dataset's balanced level of detail. 
Fact descriptions average around 652 Chinese characters, providing a succinct yet informative account of case details. 
On a per-document basis, the inclusion of roughly 1.26 charges and 4.31 statutory articles emphasizes each case's multifaceted legal context. 
Moreover, the dataset is supplemented by two extensive external corpora: one containing 103,251 judgment documents and another comprising 55,348 statutory articles. 
These resources provide rich legal background information that is particularly valuable for advanced IR technologies like retrieval-augmented generation approaches.

\subsection{Ethical Considerations}
Throughout the construction of this benchmark, we have consistently prioritized ethical considerations and taken essential measures to mitigate potential issues.
To protect sensitive information, all dataset entries have undergone strict anonymization, eliminating any personally identifiable data. 
Furthermore, to enhance transparency and facilitate reproducibility, we have publicly released the JuDGE dataset, along with all the associated models and code, on our official GitHub repository. 
This allows researchers to independently verify our work, replicate experiments, and extend our contributions.
In addition, JuDGE incorporates a Legal Corpus comprising statutes, judicial interpretations, and other authoritative legal texts. 
Given the evolving nature of legal frameworks, we are committed to maintaining the dataset’s accuracy through regular updates that reflect legislative changes and judicial developments.
Finally, to ensure broad accessibility and foster further research, JuDGE is distributed under the MIT license, which grants researchers and developers unrestricted usage rights.

% Throughout the construction of this benchmark, we have consistently prioritized ethical considerations and taken every possible measure to mitigate potential issues.
% Considering the need to protect privacy and ensure data security in legal documents, we rigorously anonymized all dataset entries. To promote transparency and facilitate future research, we have made the JuDGE dataset, along with all associated models and code, publicly available on our official GitHub repository. This commitment to openness enables other researchers to verify, replicate, and build upon our work.
% In addition, the dataset includes a carefully curated Legal Corpus, which consists of statutes, judicial interpretations, and other authoritative legal texts. Recognizing that legal statutes evolve over time, we pledge to update the JuDGE dataset regularly to reflect the most current legal information. This ensures the dataset remains accurate and relevant for ongoing research in legal informatics.
% Lastly, the JuDGE dataset is freely available under the MIT license, allowing researchers and developers to access and use the data without restrictions. 
% All related code, including the baseline implementation and the Multi-Source RAG framework for training and inference, has also been released to facilitate the reproducibility of our work.

% -------------------------------------------------

\section{Evaluation Framework}
\label{sec:evaluation}

Developing an effective evaluation framework for automatic judgment generation requires identifying the key legal criteria that determine a judgment's accuracy and quality.
Through extensive discussions with law students, legal scholars, and judges, we identified four critical dimensions that best capture the essential requirements of a well-reasoned legal judgment: 
\emph{penalty accuracy}, 
\emph{convicting accuracy}, 
\emph{referencing accuracy}, 
and \emph{documenting similarity}.
Our framework evaluates each generated judgment by comparing it to an authoritative ground-truth judgment document, ensuring alignment with actual case outcomes and established legal standards. 
In the following subsections, we first analyze the evaluation criteria for each aspect and then define specific metrics to quantify performance based on this analysis.

\subsection{Penalty Accuracy.}
In criminal proceedings, determining penalties such as prison sentences and fines is both legally important and ethically sensitive.
Even minor inaccuracies in judgment can result in procedural errors or unjust outcomes. 
To assess the precision of the penalties, we compare each predicted penalty component with its corresponding ground truth, aiming to quantitatively evaluate the document generator's performance in this aspect.

\paragraph{Metric Definition.}
To quantify how closely a predicted penalty aligns with the ground truth, we define a \emph{normalized absolute difference} for each penalty component:
\begin{equation}
    x = \frac{\left|L_{\text{pred}} - L_{\text{true}}\right|}{\max\{L_{\text{pred}}, L_{\text{true}}\}},
    % x = \frac{\left|L_{\text{pred}} - L_{\text{true}}\right|}{L_{\text{true}}},
\end{equation}
where \( L_{\text{pred}} \) and \( L_{\text{true}} \) are the predicted and ground-truth values, respectively. We then convert this difference into a score:
\begin{equation}
    S = 1 - x.
\end{equation}
% This formulation ensures that any prediction deviating by more than 100\% from the actual penalty receives a score of zero, thereby severely penalizing large errors and rewarding outputs that remain close to the true sentencing range.

\noindent This formulation ensures that \(S\) remains in the range \([0,1]\) and provides a symmetric evaluation of prediction errors.

\subsection{Convicting Accuracy.}

Accurately identifying the charges is crucial in criminal cases to ensuring that the judgment reflects the full scope of the offenses. 
Incorrect classification or omission of charges can lead to legal inaccuracies and undermine the fairness of the judgment.
To assess the accuracy of charge classification, we compare the predicted charges in the generated judgment with the charges in the authoritative ground-truth judgment. 
The goal is to evaluate the model's ability to identify all relevant charges correctly and comprehensively in the case.

\paragraph{Metric Definition.}

We measure charge-level performance using three standard classification metrics:
\begin{itemize}[leftmargin=*]
    \item \textbf{Recall}: The proportion of actual charges that the system correctly identifies.
    \item \textbf{Precision}: The proportion of predicted charges that are accurate.
    \item \textbf{F1 Score}: The harmonic mean of Precision and Recall, striking a balance between completeness and correctness.
\end{itemize}
By capturing both \emph{completeness} (Recall) and \emph{exactness} (Precision), these metrics provide a robust view of how well a system handles diverse charges.

\subsection{Referencing Accuracy.}

In jurisdictions following the civil law tradition, such as Germany, Japan, and China, the accuracy of statutory citations is fundamental to the legal validity of a ruling. Unlike common law systems, where judicial precedents play a dominant role, civil law systems rely on codified statutes as the primary legal authority. As a result, judgments must correctly and comprehensively reference the relevant legal provisions to ensure their legitimacy and adherence to statutory law. Given that this study focuses on the Chinese legal framework, which exemplifies the civil law system, an evaluation metric is introduced to assess the correctness of statutory citations in generated judgments.

\paragraph{Metric Definition.}

To measure the accuracy of statutory citations, the legal provisions cited in the generated judgment are systematically compared with those in the ground-truth judgment. Errors in statutory references may occur in two forms: under-citation, where essential legal provisions are omitted, and over-citation, where irrelevant or incorrect statutes are referenced. To assess these errors quantitatively, three standard classification metrics are employed:

\begin{itemize}[leftmargin=*]
    \item \textbf{Recall}: The proportion of correctly cited ground-truth statutes among all relevant statutes. 
    A higher Recall indicates more comprehensive legal referencing and fewer omissions.
    \item \textbf{Precision}: The proportion of correctly cited statutes among all citations in the generated judgment. 
    A higher Precision score reflects the system’s ability to avoid extraneous or incorrect legal references.
    \item \textbf{F1 Score}: The harmonic mean of Precision and Recall, balancing the completeness of legal references with their correctness.
\end{itemize}

These metrics collectively provide a rigorous evaluation of the system’s ability to identify and accurately apply statutory provisions in the generated judgments.
Since the JuDGE dataset only contains judgment documents related to criminal law, our proposed metrics are currently focused on evaluating reference accuracy in criminal law articles.

\subsection{Documenting Similarity to Ground Truth.}
Beyond evaluating specific elements such as penalties, charges, and legal references, it is essential to assess the semantic consistency of a generated judgment with the ground truth. 
Legal decisions involve complex reasoning, where the Judicial Reasoning and Judgment Result sections must align substantively with the ground truth documents.
Traditional lexical-based evaluations often fail to capture these deeper similarities, as legal reasoning can be conveyed in different but equally valid ways. 
To address this, our evaluation focuses on semantic alignment, ensuring that key arguments and legal justifications are preserved even when expressed with varying wording.

\paragraph{Metric Definition.}
We compare the {Judicial Reasoning} and {Judgment Result} sections of the generated judgment document against the ground truth document using the following semantics-based metrics:
\begin{itemize}[leftmargin=1.5em]
    \item \textbf{METEOR}~\cite{banerjee2005meteor}: Captures semantic variations and paraphrases, making it well-suited for longer and nuanced legal documents where the same reasoning may be expressed in multiple, equally valid ways.
    \item \textbf{BERTScore}~\cite{zhang2019bertscore}: Leverages contextualized embeddings to measure deep semantic alignment between generated and ground-truth texts, ensuring that subtle differences in vocabulary do not mask or inflate true alignment.
\end{itemize}
While METEOR and BERTScore are not designed specifically for legal reasoning, they offer a practical proxy for evaluating semantic consistency when domain-specific annotations are unavailable. To enhance interpretability and demonstrate the qualitative effectiveness of these metrics, we provide representative case studies on the official dataset website\footnote{https://github.com/oneal2000/JuDGE}.

% \footnote{The intuitive significance of the numerical values presented in “MET.” and “BERTS.” is further clarified by a series of case studies, which can be accessed on our dataset’s official website\url{https://github.com/oneal2000/JuDGE}}

\subsection{Automatic Evaluation Implementation}

As described in the previous subsections, our evaluation framework focuses on assessing penalties, charges, legal citations, and semantic coherence in automatically generated legal documents. 
These elements are not explicitly provided as separate outputs in the generation task (see Section~\ref{sec:task_definition}). 
Instead, the models produce complete legal documents, thus the automatic evaluation framework must automatically extract the critical features required for our metrics, such as charges, sentence lengths, fines, and cited statutes. Below, we outline the methodology and rationale behind our automatic extraction process.

Chinese legal judgments typically follow a highly standardized structure mandated by the Supreme People’s Court, which enables reliable extraction of relevant information. Each document contains three main sections: Fact Description, Judicial Reasoning, and Judgment Result. These sections are introduced by fixed phrases, such as “After trial, it was found that…” (for facts), “The court holds that…” (for reasoning), and “According to [Law], the judgment is as follows…” (for results). 
We automatically parse the generated text using these recurring lexical markers to isolate each section. 
There are various ways these sections can be phrased across different cases. Through extensive research and analysis, we have identified a wide range of alternative expressions that fulfill the same structural role in each section. 
Similarly, for the Judgment Result section, we also design and apply comprehensive regular-expression patterns to extract key features, including specific charges, sentences, fines, and cited legal articles.

Our extraction approach aligns with official guidelines and well-established templates used in Chinese courts. Based on a large-scale sampling of publicly available judgments, we verified that our rules correctly handle over 99\% of standard-format documents. 
\textbf{Consequently, if a system-generated judgment document cannot be correctly parsed by our automatic evaluation framework, this suggests a formatting error in this document.
In such cases, we set the corresponding fields (e.g., charges, sentences) to empty or zero.} 
This design choice ensures that documents that do not conform to real-world standards receive a fair penalty in our evaluation, as they do not fulfill the basic structural requirements expected in a valid judgment document.

Although the development and implementation of robust extraction rules is crucial for the accuracy of our evaluation framework, it primarily involves engineering-specific tasks. 
Since the implementation does not directly address the core scientific questions relevant to the IR and NLP community, we do not elaborate on the technical details in this paper. 
For readers interested in implementation specifics, including all regular expression patterns and parsing scripts, these are publicly available in our GitHub repository.

\section{Experimental Setup}
In this section, we present the experimental setup on the JuDGE Benchmark. Section \S \ref{sec:baselines} covers the implementation details of our baselines. In Section \S \ref{sec:llm}, we introduce the LLMs selected for this task. Finally, Section \S \ref{sec:imple} provides a detailed explanation of the implementation, including how we train and fine-tune both the retriever components and the LLMs.

\subsection{Baselines}
\label{sec:baselines}

To support future research, we systematically evaluate various baselines on our proposed JuDGE benchmark using different LLMs. 
These baselines include few-shot in-context learning and fine-tuning techniques to adapt general LLMs to the judgment document generation task. 
Additionally, we introduce an advanced baseline, Multi-source RAG (MRAG), which integrates knowledge from both statute and judgment document corpora, establishing a strong baseline performance for future reference.

\subsubsection{Few-shot In-context Learning}

In this baseline, we leverage a few representative examples from the training set as in-context demonstrations. 
Each prompt concatenates a brief instruction with exemplar pairs comprising a fact description and its corresponding full judgment document to illustrate the expected document structure and generation process. 
The LLM is then prompted to generate a judgment document for a new fact based on the pattern provided in the examples  
\footnote{Due to space constraints, we have not included the specific prompts in the main text. However, all prompts used in this study are available at the following link: \url{https://github.com/oneal2000/JuDGE}}.

\subsubsection{Supervised Fine-tuning}

For supervised fine-tuning, we directly optimize the model on the train set. 
Given a fact description \(f\) and its corresponding judgment document \(j\), the model is trained to maximize the likelihood of generating \(j\) conditioned on \(f\). 
Formally, for a document of length \(T\), the training objective is defined as:

\begin{equation}
    \mathcal{L}_{\text{FT}} = -\sum_{t=1}^{T} \log P(w_t \mid w_{<t}, f),
\end{equation}

\noindent where \(w_t\) is the token at position \(t\) and \(w_{<t}\) denotes the preceding tokens. This loss encourages the model to produce structurally coherent and legally valid judgment documents.

\subsubsection{Retrieval Augmented Generation}
In recent years, Retrieval-Augmented Generation (RAG) has emerged as a key approach to mitigating hallucinations~\cite{manakul2023selfcheckgpt,su2024mitigating,su2024unsupervised} in LLMs and enhancing their performance on knowledge-intensive tasks~\cite{borgeaud2022improving,lewis2020retrieval,su2025parametric,wang2024knowledge,guu2020retrieval,tu2025rbft,dong2025decoupling,su2024dragin}. 
The RAG paradigm follows the "Retrieval-then-Read" framework, where a retriever~\cite{zhai2008statistical,robertson2009probabilistic,zhan2021optimizing} or a complex retrieval system~\cite{salemi2024towards,chen2022web} is adopted to search for relevant information, the retrieved information is then incorporated into the input context of an LLM, enabling it to generate responses based on external knowledge. 

Building upon the RAG paradigm, we explore and propose an RAG-based approach as a baseline for our task.
In practical legal applications, generating a complete judgment document requires the integration of extensive legal information, including relevant statutory regulations, historical case precedents, and fundamental legal principles. 
To address this requirement, we propose the Multi-Source Retrieval-Augmented Generation (MRAG) baseline that integrates external knowledge to enhance the generation process. 
This baseline divides judgment generation into two phases: Information Collection and Document Generation. 
In the Information Collection phase, two distinct retrievers are employed to retrieve relevant information from different sources. 
Specifically, the Law Retriever targets relevant statutory regulations, while the Case Retriever focuses on relevant judgment documents. 
In the Document Generation phase, the LLM leverages the fact description along with the information gathered in the information collection phase to generate the complete judgment document.

The Law Retriever is designed to retrieve relevant statutes based on a case’s fact description. 
It employs a dual-encoder architecture, where the relevance score between the factual description \(f\) and a statute \(s\) is defined as the dot product of their respective embeddings. 
Formally, we define:

\vspace{-3mm}

\begin{equation}
\label{embedding}
Emb(X) =  transformer_{[CLS]}(X) ,
\end{equation}

\begin{equation}
\label{score_dense}
\begin{split}
S(q, s) = Emb(f)^\top \cdot Emb(s) ,
\end{split}
\end{equation}

\noindent where \(f\) denotes the factual description and \(s\) denotes the statute. 
The function \(\mathrm{transformer}_{[CLS]}(\cdot)\) produces a contextualized vector for each token, and we select the \([CLS]\) token’s vector as the input’s embedding. 
In Equation~\ref{score_dense}, the dot product of the embeddings is used as the relevance score \(S\). 
To train the Law Retriever, we employ a contrastive learning strategy. Specifically, for each instance in the JuDGE dataset, the model is trained to retrieve law articles relevant to a given Fact. 
The law articles cited in the ground truth judgment document serve as positive examples, while all other articles in the corpus that are not cited are treated as negative examples.  
During training, each fact is paired with its corresponding positive law article \(a_i^+\). To construct the negative set, we randomly sample six negative articles from the statute corpus, enabling the model to distinguish relevant provisions from irrelevant ones. The model is then optimized using the following loss function:  

\begin{equation}
\mathcal{L}(f,a_i^+,N) = -\log \frac{\exp(S(f,a_i^+))}{\exp(S(f,a_i^+)) + \sum_{a^- \in N} \exp(S(f,a^-))},
\end{equation}

\noindent where \(N\) represents the set of irrelevant law articles, and $S$ is defined in Equation ~\ref{score_dense} . This objective encourages the model to assign higher relevance scores to the relevant law articles while minimizing the scores of non-relevant ones.

\begin{table}
\centering
\setlength\tabcolsep{3pt}
\caption{Comparison of the Law Retriever in the MRAG baseline with traditional retrieval algorithms. Metrics include MRR, Precision (P), and Recall (R) at various cutoffs (5 and 10). The best results are in bold.}
\label{tab:retrieve}
\begin{tabular}{cccccc}
\toprule
                       & \textbf{MRR@100}& \textbf{P@5}    & \textbf{P@10}   & \textbf{R@5}    & \textbf{R@10}    \\
                       \midrule
\textbf{TF-IDF}        & 0.3605          & 0.1625          & 0.1200          & 0.2096          & 0.3074           \\
\textbf{BM25}          & 0.4373          & 0.1501          & 0.1024          & 0.1958          & 0.2649           \\
\textbf{Law Retriever} & \textbf{0.8328} & \textbf{0.4535} & \textbf{0.3509} & \textbf{0.5529} & \textbf{0.8262} \\
% \textbf{Law Re-ranker} & \textbf{0.9422} & \textbf{0.6048} & \textbf{0.3810} & \textbf{0.7438} & \textbf{0.9038} \\
\toprule
\end{tabular}
\end{table}

In Table~\ref{tab:retrieve}, we present the performance of our Law Retriever on the statute retrieval task and compare it with traditional lexical-based methods, TF-IDF~\cite{ramos2003using} and BM25~\cite{robertson2009probabilistic}. 
The results show that our Law Retriever significantly outperforms these baselines across all reported metrics, highlighting its ability to capture deeper semantic relationships between the factual descriptions and the statutory texts. 
Given that relevant legal provisions are often not lexically identical to the facts, relying solely on exact term matching proves insufficient.
Consequently, these findings underscore the importance of training a specialized dense retrieval model.

For the Case Retriever, we employ the pre-trained dense retrieval model SAILER~\cite{li2023sailer}, a structure-aware language model specifically designed for legal document representation. 
The retrieval process follows the standard approach used in the Legal Case Retrieval task~\cite{ma2021lecard,su2024pre,ma2023caseencoder,li2023thuir}, a well-established and extensively studied problem in the information retrieval community.
Given that SAILER has consistently demonstrated strong performance in this domain, we adopt it directly as our retriever and do not further elaborate on the retrieval process in this paper.

Following the retrieval stage, the base language model is fine-tuned to generate the final judgment document. This fine-tuning is conducted using a prompt template that concatenates the case fact description with the top retrieved legal information—specifically, the two most relevant cases from the Case Retriever and the ten most pertinent statutes from the Law Retriever. The language model is trained using a next-token prediction loss defined as:

\begin{equation}
\mathcal{L}_{\text{FT}} = -\sum_{t=1}^{T} \log P(w_t \mid w_{<t}, \text{context}),
\end{equation}

\noindent where \(T\) is the length of the judgment document, and the context comprises both the fact description and the retrieved legal knowledge. During inference, the same prompt structure is employed, with dynamically retrieved cases and statutes replacing their training-time counterparts. This retrieval-augmented strategy enables the language model to generate judgment documents that are not only structurally complete but also legally coherent and grounded in external legal information.

\begin{table*}
\centering
\setlength\tabcolsep{3pt}
\caption{Experimental results on multiple baselines across six different LLMs. QW-3B-Base refers to the Qwen-2.5-3B model, while QW-3B-Chat refers to the Qwen-2.5-3B-Instruct model (the same naming convention applies to the 7B models). “MET.” denotes METEOR, “BERTS.” for BERTScore, and “Prec.” for precision. The best results for each LLM are highlighted in bold.}
\vspace{-2mm}
\label{tab:main}
\begin{tabular}{cccccccccccccc}
 \toprule
                    &                   & \multicolumn{2}{c}{\textbf{Penalty Acc.}} & \multicolumn{3}{c}{\textbf{Convicting Acc.}}          & \multicolumn{3}{c}{\textbf{Referencing Acc.}} & \multicolumn{2}{c}{\textbf{Reasoning Section}} & \multicolumn{2}{c}{\textbf{Judgment Section}}  \\
\cmidrule(lr){3-4}\cmidrule(lr){5-7}\cmidrule(lr){8-10}\cmidrule(lr){11-12}\cmidrule(lr){13-14}
\textbf{Model}      & \textbf{Method}   & \textbf{Prison} & \textbf{Fine}       & \textbf{Recall} & \textbf{Prec.} & \textbf{F1} & \textbf{Recall} & \textbf{Prec.} & \textbf{F1} & \textbf{MET.} & \textbf{BERTS.}                  & \textbf{MET.} & \textbf{BERTS.}                  \\
 \toprule
\multirow{2}{*}{\textbf{QW-3B-Base}} & \textbf{SFT} & 0.5975 & \textbf{0.5149} & 0.9381 & 0.9375 & 0.9378 & 0.6915 & 0.6441 & 0.6669 & \textbf{0.6281}& 0.8400& \textbf{0.7147}& 0.7408
\\
                                   & \textbf{RAG} & \textbf{0.6273} & 0.5132 & \textbf{0.9471} & \textbf{0.9511} & \textbf{0.9491} & \textbf{0.7569} & \textbf{0.6847} & \textbf{0.7190} & 0.5945& \textbf{0.8412}& 0.7045& \textbf{0.8307}\\
\midrule
\multirow{2}{*}{\textbf{QW-7B-Base}}  & \textbf{SFT} & 0.6380 & 0.5273 & \textbf{0.9651} & \textbf{0.9664} & \textbf{0.9657} & \textbf{0.7500} & 0.7187 & 0.7340 & \textbf{0.6523}& 0.8527& \textbf{0.7548}& 0.8025
\\
                                   & \textbf{RAG} & \textbf{0.6489} & \textbf{0.5458} & 0.9411 & 0.9441 & 0.9426 & 0.7356 & \textbf{0.7649} & \textbf{0.7502} & 0.6020& \textbf{0.8533}& 0.7047& \textbf{0.8565}\\
                                   \midrule
\multirow{3}{*}{\textbf{LexiLaw-6B}}   & \textbf{Direct}& 0.0408& 0.0261& 0.6178& 0.6228& 0.6202& 0.0010& 0.0040& 0.0016
& 0.0994& 0.7259& 0.3191& 0.0043
\\
                                   & \textbf{ICL}& 0.0102& 0.0126& 0.4072& 0.4112& 0.4092& 0.0071& 0.0088& 0.0078
& 0.3403& 0.7179& 0.1636& 0.0058
\\
                                   & \textbf{SFT}& \textbf{0.5926}& \textbf{0.4942}& \textbf{0.9401}& \textbf{0.9288}& \textbf{0.9344}& \textbf{0.6030}& \textbf{0.6908}& \textbf{0.6439}
& \textbf{0.6471}& \textbf{0.8399}& \textbf{0.6907}& \textbf{0.6715}\\
\midrule
\multirow{3}{*}{\textbf{Hanfei-7B}}    & \textbf{Direct}& 0.4902& 0.3541& 0.8623& 0.8683& 0.8653& 0.4624& 0.5306& 0.4941
& 0.5132& 0.7643& 0.5433& 0.4860\\
                                   & \textbf{ICL}& 0.0635& 0.0282& 0.1128& 0.1118& 0.1123& 0.0209& 0.0407& 0.0276
& 0.3184& 0.5404& 0.2632& 0.0434
\\
                                   & \textbf{SFT}& \textbf{0.6520}& \textbf{0.5526}& \textbf{0.9381}& \textbf{0.9318}& \textbf{0.9350}& \textbf{0.7238}& \textbf{0.6813}& \textbf{0.7019}& \textbf{0.6312}& \textbf{0.8389}& \textbf{0.7141}& \textbf{0.7865}\\
\midrule
\multirow{4}{*}{\textbf{QW-3B-Chat}} & \textbf{Direct}& 0.6117 & 0.4868 & 0.8713 & 0.8752 & 0.8732 & 0.5972 & 0.7300 & 0.6570 & 0.4620& 0.7724& 0.3658& 0.7120\\
                                   & \textbf{ICL}& 0.6278 & 0.4720 & 0.8832 & 0.8746 & 0.8789 & 0.4928 & 0.7092 & 0.5815 & 0.4245& 0.6901& 0.3699& 0.6790\\
                                   & \textbf{SFT}      & \textbf{0.6673} & \textbf{0.5443} & 0.9401 & 0.9444 & 0.9423 & 0.7205 & \textbf{0.7503} & 0.7351 & \textbf{0.5836}& \textbf{0.8509}& 0.6803& \textbf{0.8889}\\
                                   & \textbf{RAG}      & 0.6527 & 0.5296 & \textbf{0.9511} & \textbf{0.9531} & \textbf{0.9521} & \textbf{0.7292} & 0.7484 & \textbf{0.7387} & 0.5807& 0.8425& \textbf{0.7095}& 0.8851
\\
\midrule
\multirow{4}{*}{\textbf{QW-7B-Chat}} & \textbf{Direct}& 0.6655 & 0.5075 & 0.9242 & 0.9242 & 0.9242 & 0.5045 & \textbf{0.8161} & 0.6235 & 0.5061& 0.8098& 0.4161& 0.7823
\\
                                   & \textbf{ICL}& 0.6731 & 0.5095 & 0.9371 & 0.9385 & 0.9378 & 0.6537 & 0.7915 & 0.7161 & 0.5436& 0.8245& 0.6220& 0.8523
\\
                                   & \textbf{SFT}      & 0.6604 & 0.5506 & 0.9521 & 0.9518 & 0.9519 & 0.7787 & 0.7694 & \textbf{0.7740} & \textbf{0.6220}& \textbf{0.8597}& \textbf{0.7242}& \textbf{0.8961}\\
                                   & \textbf{RAG}      & \textbf{0.6795} & \textbf{0.5512} & \textbf{0.9611} & \textbf{0.9604} & \textbf{0.9607} & \textbf{0.7955} & 0.7509 & 0.7726 & 0.6076& 0.8513& 0.7236& 0.8887\\

 \toprule
\end{tabular}
\vspace{-3mm}
\end{table*}

\subsection{Selected Large Language Models} \label{sec:llm}

To evaluate the performance of various LLMs on our proposed \textsc{JuDGE} dataset, we conduct experiments using both general-purpose and legal-domain models.

\subsubsection{General-Purpose Models.} 
We use four variants from the Qwen 2.5 series~\cite{qwen2025qwen25technicalreport}, consisting of both 3B and 7B parameter scales and two model types (base vs. instruct). 
Specifically, the base models, Qwen-2.5-3B and Qwen-2.5-7B are pre-trained on large-scale corpora without alignment to human instructions. 
In contrast, the instruct versions, Qwen-2.5-3B-Instruct and Qwen-2.5-7B-Instruct, are fine-tuned with instruction tuning and related alignment methods to improve their ability to follow human instructions.

\subsubsection{Legal-Domain Models.} 
We select {HanFei-1.0} (7B) and {LexiLaw} (6B) for domain-specific LLM experiments.
{HanFei-1.0}\footnote{https://github.com/siat-nlp/HanFei} is trained on a broad corpus of law-related texts, including legal news, forums, statutes, judicial interpretations, legal consultations, bar exam questions, and court judgments. This model is designed to enhance its reasoning and consultation capabilities in legal contexts.
{LexiLaw}\footnote{https://github.com/CSHaitao/LexiLaw} is built upon the ChatGLM-6B model and has been further fine-tuned on specialized legal data to improve its performance and knowledge specificity for legal tasks.

% Our selected LLMs are listed as follows:
% \begin{itemize}[leftmargin=*]

% \item \textbf{Baichuan}~\cite{yang2023baichuan} is a series of large-scale multilingual language models, trained from scratch on 2.6 trillion tokens. 
% We choose the \textbf{Baichuan-2-Base-13B} model which is widely used in bilingual Chinese-English scenarios.

% \item \textbf{ChatGLM}~\cite{du2022glm} is a series of generative language models optimized for Chinese question answering and dialogue. We choose \textbf{ChatGLM3-6B} with 6.2 billion parameters.

% \item \textbf{ChatGPT} ~\cite{brown2020language} is a series of large language models developed by OpenAI, including several versions. Among these, we choose \textbf{GPT-3.5-turbo}, which is identified as the most advanced GPT-3.5 model.

% \end{itemize}

\subsection{Implementation Details}
\label{sec:imple}

For the retrieval module in the RAG baseline, we train the dense retriever as follows.
The Law Retriever are initialized using the pre-trained Chinese-RoBERTa-WWM\footnote{https://huggingface.co/hfl/chinese-roberta-wwm-ext}.
We use the AdamW optimizer with a learning rate of $5 \times 10^{-6}$, with mixed-precision ($fp16$) training for enhanced efficiency. 
The models are trained for two epochs.
For fine-tuning the LLM, we use the Deepspeed framework with Zero-2 optimization to enable efficient full-model fine-tuning. 
This process employs the AdamW optimizer, a learning rate of $3 \times 10^{-5}$, and mixed-precision ($bf16$) training for improved memory efficiency. 
The LLM is fine-tuned for four epochs on the dataset.
All training processes are conducted on a server equipped with eight NVIDIA A100 GPUs, each with 40GB of memory. 
For the generation phase of LLMs, all experiments are conducted using the publicly available Hugging Face implementations. 
We use the default hyperparameters and chat template provided in their official Hugging Face repository.
All the prompt templates used in this paper are available in our GitHub repository\footnote{https://github.com/oneal2000/JuDGE}.

% For the retrieval module in the RAG baseline, we train the Dense Retriever and re-ranker as follows:
% The training set is randomly split into two halves. One half is used to train the Retriever, while the other half is used to train the re-ranker based on the top-100 articles returned by the Retriever.
% We initialize both the re-ranker and Retriever with the pre-trained Chinese-RoBERTa-WWM\footnote{https://huggingface.co/hfl/chinese-roberta-wwm-ext}.
% The training is performed using the AdamW optimizer with a learning rate of $5e^{-6}$, and we utilize mixed precision (fp16) training for efficiency. The model is trained for 2 epochs.
% For fine-tuning the LLM, we employ the Deepspeed framework with Zero-2 optimization to perform full model fine-tuning. The training is carried out with the AdamW optimizer, a learning rate of $3e^{-5}$, and mixed precision (bf16) training for improved memory efficiency. The LLM is fine-tuned for 4 epochs on the dataset.
% All the training process are conducted on a server equipped with 8 NVIDIA A100 GPUs, each with 40GB of memory. 
% For {generation configuration}, all experiments are conducted using the publicly released Hugging Face implementations. 
% We adopt the default hyperparameters and chat template provided in the official Huggingface repository. 
% All LLM is deployed on a server equipped with 8 NVIDIA A100 GPUs, each with 40GB of memory.

\section{Experimental Results}
\label{sec:exp_results}

In this section, we present the main experimental results on the \textsc{JuDGE} benchmark and provide an in-depth analysis of the findings. 
As shown in Table~\ref{tab:main}, we compare several baselines across different LLMs, including both base and chat-oriented models, as well as legal LLMs. 
All the prompt templates used in the experiments are detailed in our official GitHub repository\footnote{https://github.com/oneal2000/JuDGE}.
We omit direct generation and in-context learning on the two base models (QW-3B-Base and QW-7B-Base) because they are not instruction-tuned or aligned with human preferences. 
Consequently, their default behavior is to continue text generation rather than following a structured legal drafting instruction. Below, we summarize the key observations:

\textbf{(1)} Legal LLMs (e.g., \textit{LexiLaw-6B} and \textit{Hanfei-7B}) show notably poor performance when generating judgment documents directly or via few-shot in-context learning. 
A closer examination of the generated outputs reveals that these models often reuse or partially copy the exemplars provided in the prompt, neglecting to incorporate the unique facts of each case. 
We suspect that this phenomenon arises from catastrophic forgetting. 
Since these models were continually pre-trained and fine-tuned on large-scale legal data (e.g., legal QA), their parameters have diverged significantly from their initial state, resulting in the loss of previously acquired knowledge and instruction-following capabilities. 
\textbf{(2)} Few-shot in-context learning benefits larger chat models more significantly.
For QW-7B-Chat, introducing a few-shot example in the prompt consistently enhances performance across almost all the metrics. 
In contrast, QW-3B-Chat sees only marginal gains from in-context learning. 
This difference highlights that larger models with more capacity generally exhibit stronger in-context learning capabilities.
\textbf{(3)} Supervised fine-tuning substantially improves performance.
Regardless of the models' initial capability, supervised fine-tuning (SFT) leads to improvements over direct generation and in-context learning. 
This trend is observed across both general-purpose models (e.g., QW-3B-Chat and QW-7B-Chat) and legal LLMs (e.g., LexiLaw-6B and Hanfei-7B). 
Notably, the improvement is particularly large for legal LLMs: while their performance is almost unusable before fine-tuning, it becomes acceptable and sometimes competitive after the task-specific SFT. 
These findings reinforce the importance of supervised adaptation for certain downstream tasks.
\textbf{(4)} Multi-source retrieval-augmented generation (RAG) offers competitive advantages. 
For instance, MRAG consistently achieves the highest recall, precision, and F1 score in convicting accuracy. 
However, while MRAG achieves high recall in referencing accuracy, its precision is relatively lower.
We attribute this to our top-$k$ retrieval strategy (here $k=10$), which occasionally supplies multiple irrelevant statutes. 
As indicated by Table~\ref{tab:basic_statistics}, only 4.31 criminal statutes are truly relevant per case. Since current LLMs are not fully adept at filtering out misleading articles, this results in a precision drop.

\section{Related Work}
\subsection{Legal Information Retrieval}
Legal information retrieval (IR) differs from open-domain IR as it requires the retrieval model to incorporate legal knowledge and consider both semantic and legal perspectives when modeling the text. 
A prominent research direction is \emph{Legal Case Retrieval}, which focuses on retrieving past legal cases that are relevant to a given query case, aiding legal professionals in case analysis and decision-making~\cite{su2023caseformer,ma2021lecard,ma2023caseencoder}.
Early work in this area primarily relied on lexical matching methods to assess similarity~\cite{rosa2021yes}. 
Recently, with the development of {Dense Retrieval}~\cite{karpukhin2020dense,zhan2021optimizing,su2023thuir2,fang2024scaling,su2023wikiformer}, researchers have begun to explore methods to enhance retrievers with deeper legal understanding via large-scale pretraining on unlabeled legal corpora. 
For instance, Sailer~\cite{li2023sailer} trains the dense retriever through an autoencoder-like approach, where the embedding of a case’s {Fact} section is used to reconstruct other sections of the document, thereby infusing the fact embeddings with richer contextual information. 
Caseformer~\cite{su2023caseformer} proposes an unsupervised contrastive learning approach for automatically measuring similarity between legal documents. 
% The method leverages legal case attributes such as charges and cited legal provisions to automatically generate positive and negative cases relative to a query case, thereby training dense retrieval models. This fully automated process enables the generation of substantial training data from unlabeled corpora, enhancing retrieval performance.
Another major direction is \emph{Statute Retrieval}, which aims to locate and retrieve relevant statutory articles for given a query. 
For instance, the annual COLIEE competitions~\cite{goebel2023summary,kim2022coliee,rabelo2022overview,li2023thuir,li2023thuir2} focus on Japanese legal bar exam questions, where the goal is to retrieve relevant statutes from the Japanese Civil Code. 
Similarly, the AILA~\cite{bhattacharya2019fire} targets the Indian legal system, using queries from Supreme Court of India judgments to retrieve relevant statutes. 
Beyond professional use cases, statute retrieval also caters to non-professional users who may lack formal legal training; for example, STARD~\cite{su2024stard} is specifically designed to address layperson queries, providing plain-language questions paired with relevant statutory content.
Beyond retrieval, \emph{Legal QA} tasks aim to satisfy the public’s need for accessible legal information~\cite{do2017legal}, such as the French long-form QA dataset LLeQA~\cite{louis2024interpretable} and GerLayQA~\cite{buttner2024answering}, which contains laymen’s German legal questions paired with authoritative law book paragraphs.

\subsection{Legal Judgment Prediction}

Legal Judgment Prediction (LJP) is a fundamental task in many Civil Law jurisdictions. 
It aims to infer judicial outcomes based on a case's factual description, such as applicable legal articles, charges, and penalty terms. 
For instance, ~\citeauthor{luo2017learning}~\cite{luo2017learning} integrates statutory details to enhance charge classification by leveraging the inherent structure of legal codes. Similarly, ~\citeauthor{hu2018few}~\cite{hu2018few} proposes a few-shot learning approach to address data scarcity in charge prediction tasks. 
Building on these efforts, researchers introduce advanced neural architectures, including multi-channel attentive networks and gating mechanisms to model the complex interactions among legal facts, charges, and statutes~\cite{li2019mann,kang2019creating,chen2019charge}.
More recently, researchers have explored the applicability of large language models for LJP, showing that LLMs can further enhance predictive accuracy under realistic conditions~\cite{nigam2024rethinkinglegaljudgementprediction}.

Although previous efforts have notably advanced the accuracy of LJP, they typically formulate the task as a classification problem (i.e., predicting discrete labels for charges, articles, and sentences). 
In contrast, our work focuses on the \emph{generative} task, requiring the model to produce the \textbf{complete judgment document} based on the facts and relevant legal knowledge. 
This extends beyond merely outputting labels: the generation process must logically organize all necessary legal elements into a full judgment document. 
Such an approach more closely aligns with practical legal scenarios, where judges produce well-structured rulings that explicitly cite relevant statutes, articulate charges, and determine sentencing outcomes. 
As a result, the JuDGE benchmark broadens the scope of LJP from classification to generation and bridges the gap between legal classification tasks and real-world judgment document drafting.

\section{Conclusion}

In this paper, we introduced JuDGE, a benchmark for evaluating the generation of judgment documents. 
By formalizing the task and providing the benchmark dataset, we establish a solid foundation for further research in this important yet underexplored area. 
Through collaboration with legal professionals, we also developed an automated evaluation framework that measures performance across four key dimensions: penalty accuracy, convicting accuracy, referencing accuracy, and similarity to ground truth.
Our experimental results reveal that current approaches struggle to produce high-quality judgment documents. 
Although supervised fine-tuning and our proposed Multi-source RAG approach show improvements, the performance gap indicates that judgment document generation still remains challenging and requires further research and innovation.
We hope JuDGE will inspire more advanced techniques in judgment document generation, ultimately helping the legal community improve efficiency and reduce manual workloads.

\bibliographystyle{ACM-Reference-Format}
% \balance
\bibliography{sample-base}

\end{document}